\documentclass[letterpaper, 10 pt, conference]{ieeeconf}  

\IEEEoverridecommandlockouts                              

\usepackage{cite}
\usepackage{graphics} 
\usepackage{epsfig} 
\usepackage{mathptmx} 
\usepackage{times} 
\usepackage{amsmath} 
\usepackage{amssymb}  
\usepackage[linesnumbered,ruled,vlined]{algorithm2e}
\usepackage{algorithmic}
\usepackage[font=footnotesize,labelfont=footnotesize]{caption}
\usepackage[font=footnotesize,labelfont=footnotesize]{subcaption}
\usepackage{xcolor}
\usepackage{bm}
\usepackage{hyperref}
\usepackage{newtxtext, newtxmath}

\DeclareMathOperator*{\argmin}{arg\,min}
\DeclareMathOperator*{\argmax}{arg\,max}

\newtheorem{corollary}{Corollary}

\newtheorem{definition}{Definition}

\title{\LARGE \bf
Online Connectivity-aware Dynamic Deployment for Heterogeneous Multi-Robot Systems }

\author{Chendi Lin$^{1}$, Wenhao Luo$^{1}$, and Katia Sycara$^{1}$
\thanks{$^{1}$The authors are with the Robotics Institute, School of Computer Science, Carnegie Mellon
University, Pittsburgh, PA 15213, USA. 
        Email: {\tt\small chendil@alumini.cmu.edu,\{wenhaol,katia\}@cs.cmu.edu}}%
}

\begin{document}

\maketitle
\thispagestyle{empty}
\pagestyle{empty}

\begin{abstract}
In this paper, we consider the dynamic multi-robot distribution problem where a heterogeneous group of networked robots is tasked to spread out and simultaneously move towards multiple moving task areas while maintaining connectivity. 
The heterogeneity of the system is characterized by various categories of units and each robot carries different numbers of units per category representing heterogeneous capabilities. Every task area with different importance demands a total number of units contributed by all of the robots within its area. Moreover, we assume the importance and the total number of units requested from each task area is initially unknown. The robots need first to explore, i.e., reach those areas, and then be allocated to the tasks so to fulfill the requirements. The multi-robot distribution problem is formulated as designing controllers to distribute the robots that maximize the overall task fulfillment while minimizing the traveling costs in presence of connectivity constraints. 
We propose a novel connectivity-aware multi-robot redistribution approach that accounts for dynamic task allocation and connectivity maintenance for a heterogeneous robot team. 
Such an approach could generate sub-optimal robot controllers so that the amount of total unfulfilled requirements of the tasks weighted by their importance is minimized and robots stay connected at all times. Simulation and numerical results are provided to demonstrate the effectiveness of the proposed approaches. 
\end{abstract}

\section{INTRODUCTION}
\label{sec:intro}

\begin{figure*}[h]
\centering
\begin{subfigure}{0.25\textwidth}
\includegraphics[width=0.8\textwidth]{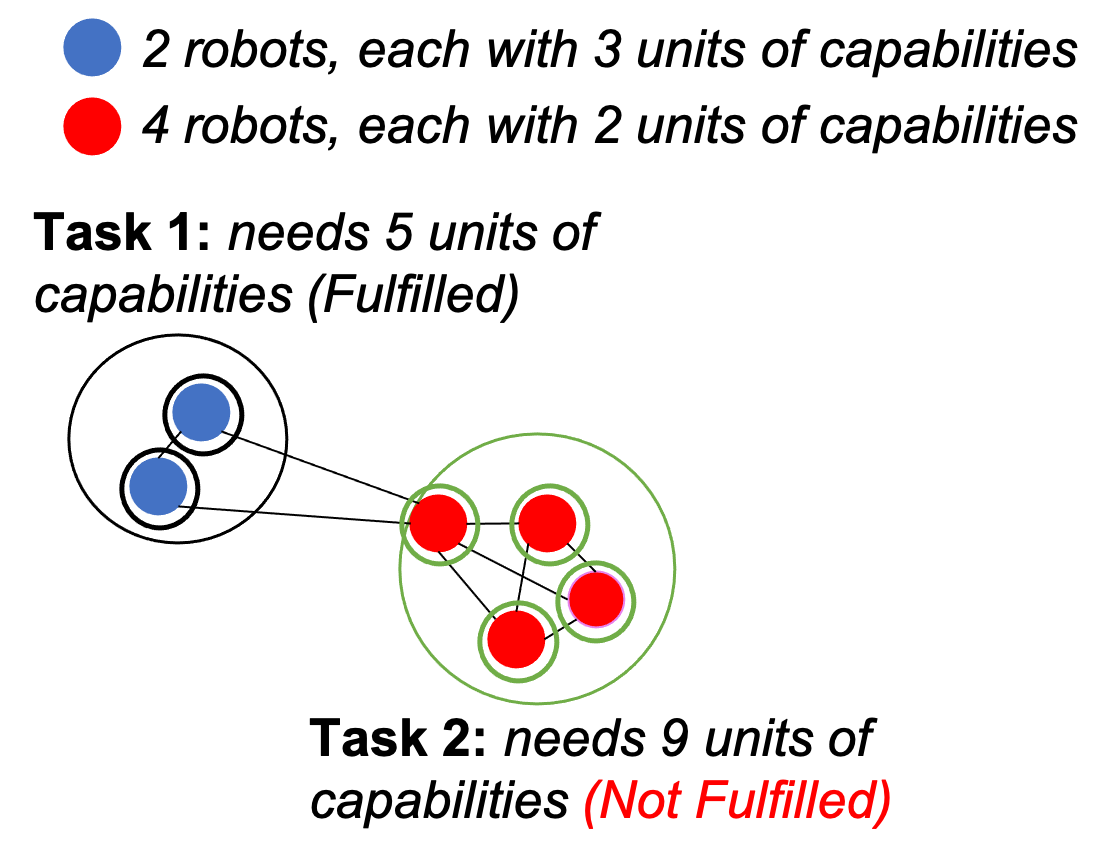}
\caption{Task 1 is fulfilled but Task 2 is not.}
\label{fig:combo1}
\end{subfigure} \hfill
\begin{subfigure}{0.25\textwidth}
\includegraphics[width=0.8\textwidth]{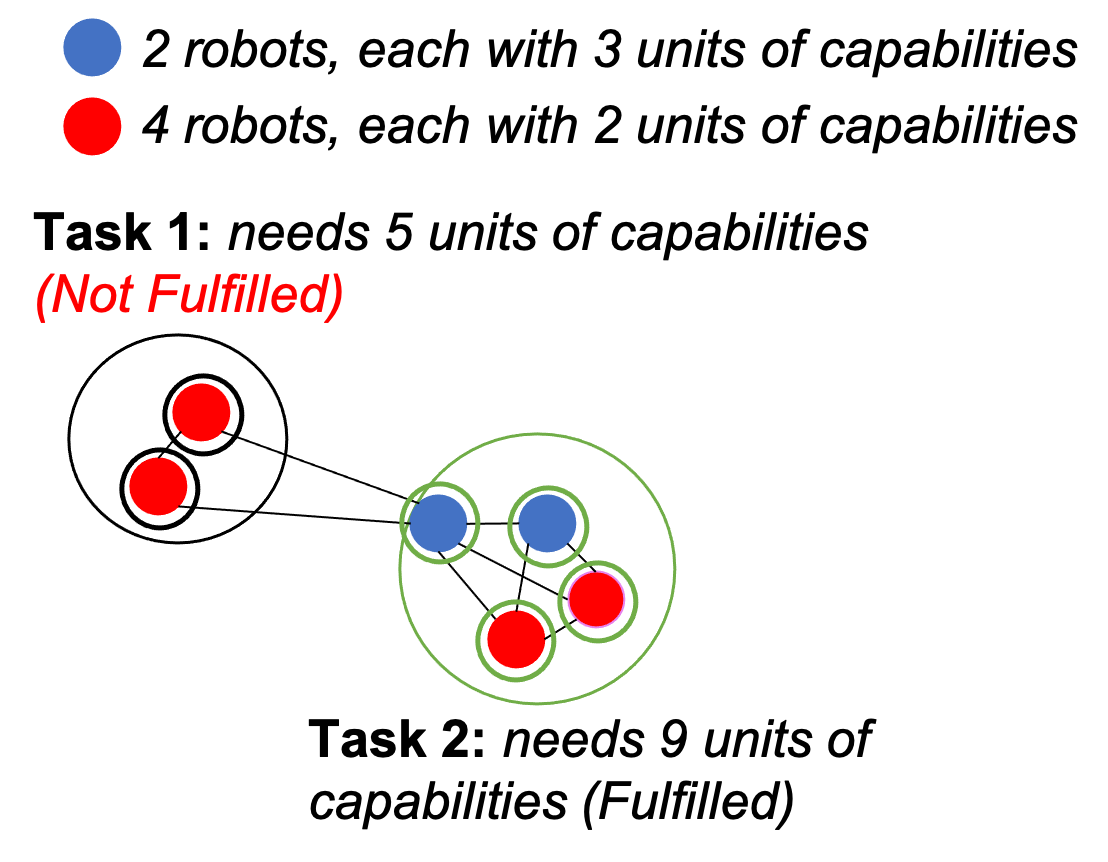}
\caption{Task 2 is fulfilled but Task 1 is not.}
\label{fig:combo2}
\end{subfigure} \hfill
\begin{subfigure}{0.25\textwidth}
\includegraphics[width=0.8\textwidth]{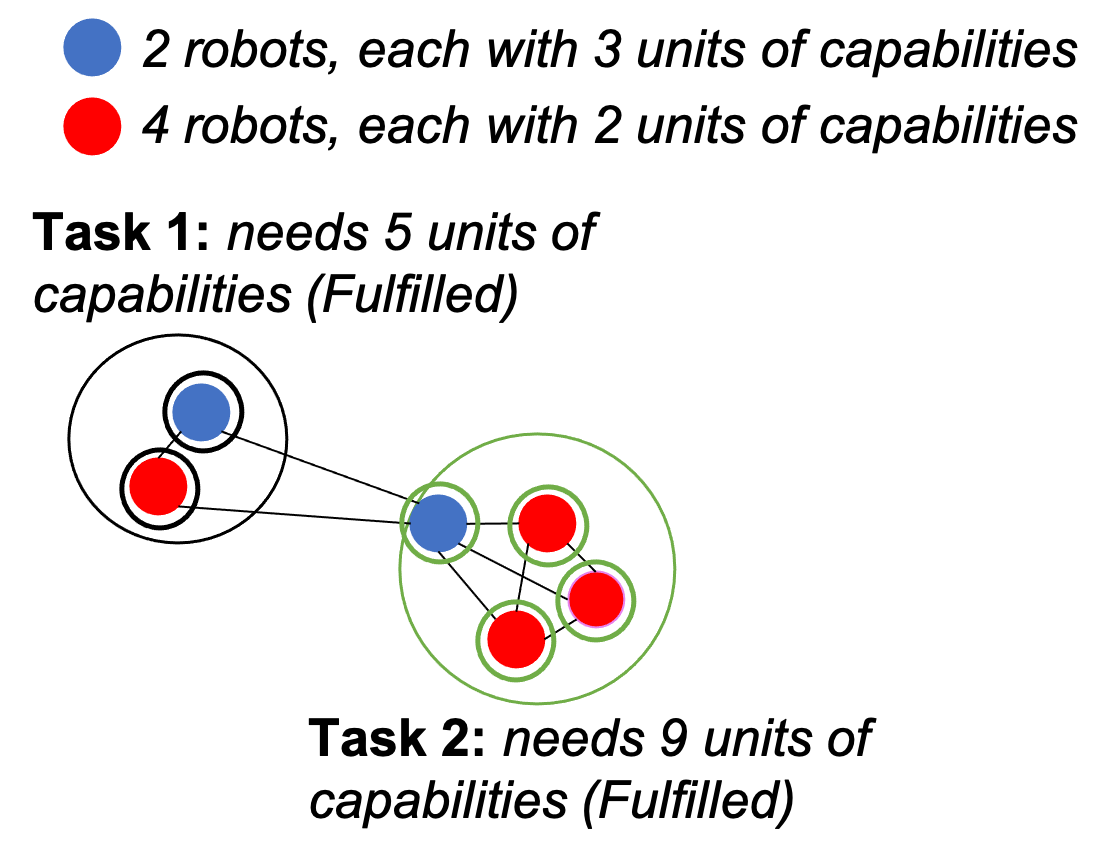}
\caption{Both tasks are fulfilled.}
\label{fig:combo3}
\end{subfigure} \hfill
\caption{Illustration of combinatorial task allocation that two types of robots with one category of capability allocated to two different tasks: 2 blue robots with 3 units of capabilities each and 4 red robots with 2 units of capabilities each.  Task 1 requires 5 units of capabilities, and Task 2 needs 9 units. With the assignment in (a), Task 1 is over-allocated of 1 extra unit by two blue robots while Task 2 misses 1 unit. In (b) Task 2 is over-allocated whereas Task 1 is under-allocated by 1 unit. The combination shown in (c) can fully meet the requirements of both tasks. }
\label{fig:combinatorial}
\end{figure*}

Multi-robot systems are powerful in their ability to perform different tasks in parallel. 
Typically, the robots are distributed to different tasks based on the their capabilities, the importance, and the needs of the tasks. One application is the DARPA Offset Projects, where the UAVs and UGVs coordinate to explore and execute the tasks in both the indoor and outdoor environments. The allocation of the robots will depend on their capabilities like sensors and mobility, the needs and the priority of the tasks.
For example, a task may involve a target that needs 8 units of load-carrying capability. It can be supplied by 4 robots, each of which has 2 units of capability, or by 3 robots, where two  have 3 units of capability each and one that has 2 units of capability. Moreover, one task may involve a target where the robots would be concentrated in the middle of the target for better monitoring, whereas another task may require the robots to encircle the target. Therefore, different controllers would also be needed.
In this work, we discuss our approach using the illustrative example of covering targets that may appear dynamically. However, the method applies to multiple other  tasks.
Figure \ref{fig:combinatorial} shows a simple example of such problems with one category of capability. In this example, two types of robots are allocated to two tasks with known locations, areas, and needs. We mark robots of different types by different colors. Allocating more than the required capabilities (over-allocating) does not bring any extra utility. 
Figure \ref{fig:combo1} and Fig. \ref{fig:combo2} lead to a lack of 1 unit in one of the tasks whereas the allocation in Fig. \ref{fig:combo3} fulfills the requirements of both tasks. 

\begin{figure*}[t]
\centering
\begin{subfigure}{0.3\textwidth}
\includegraphics[width=0.8\textwidth]{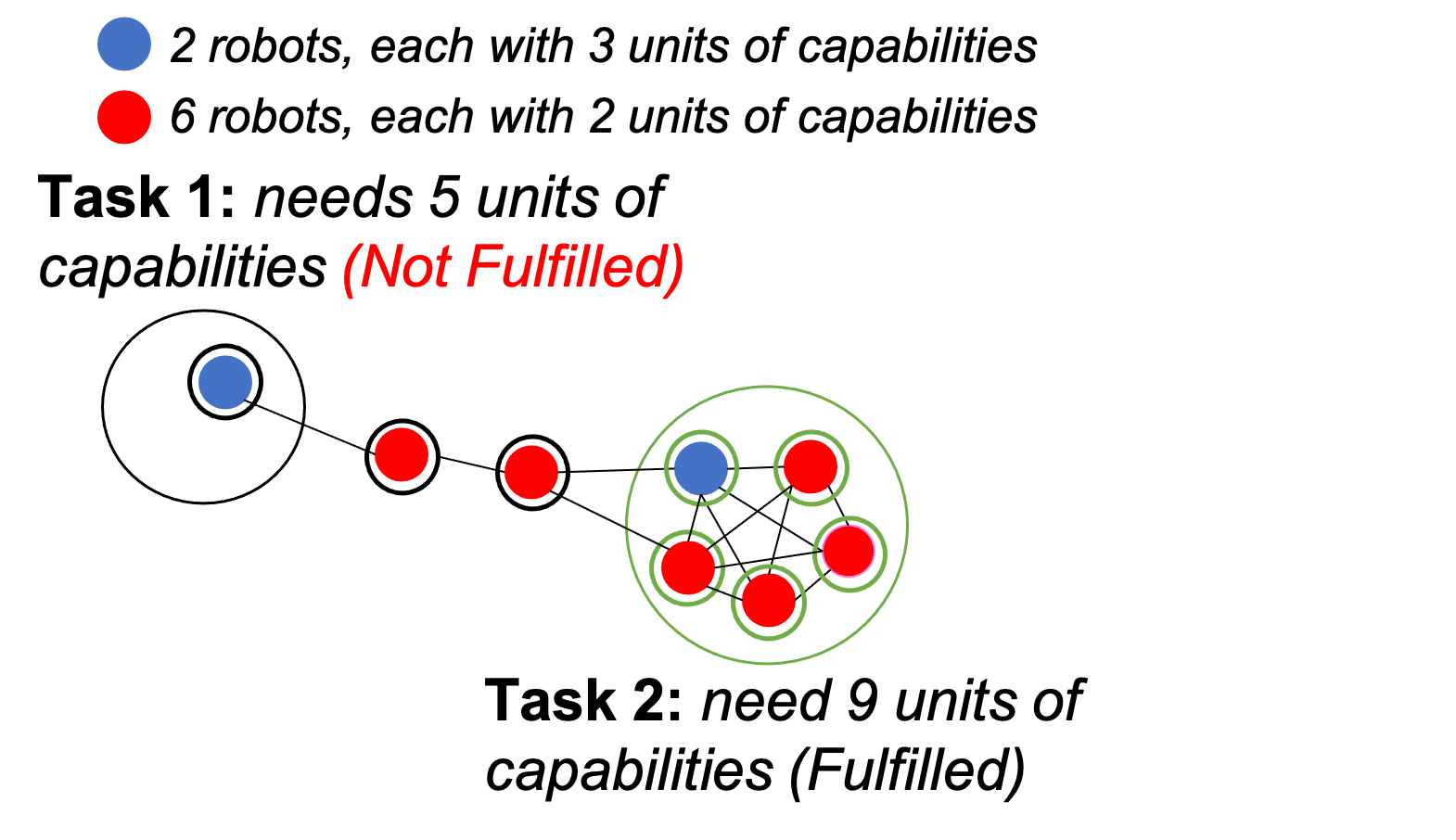}
\caption{Task 1 is fulfilled but Task 2 is not.}
\label{fig:introduction1}
\end{subfigure} \hfill
\begin{subfigure}{0.25\textwidth}
\includegraphics[width=0.8\textwidth]{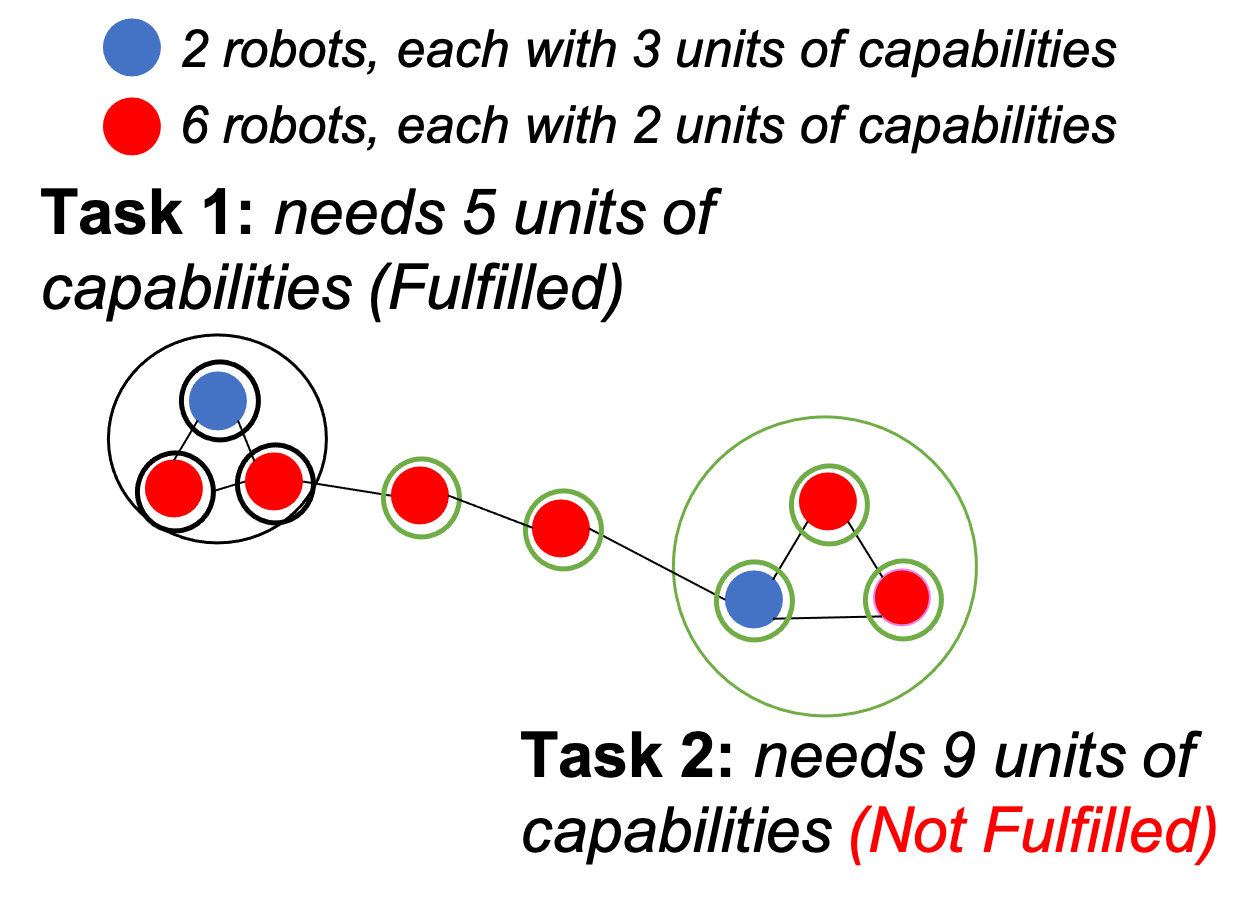}
\caption{Task 2 is fulfilled but Task 1 is not.}
\label{fig:introduction2}
\end{subfigure} \hfill
\begin{subfigure}{0.25\textwidth}
\includegraphics[width=0.8\textwidth]{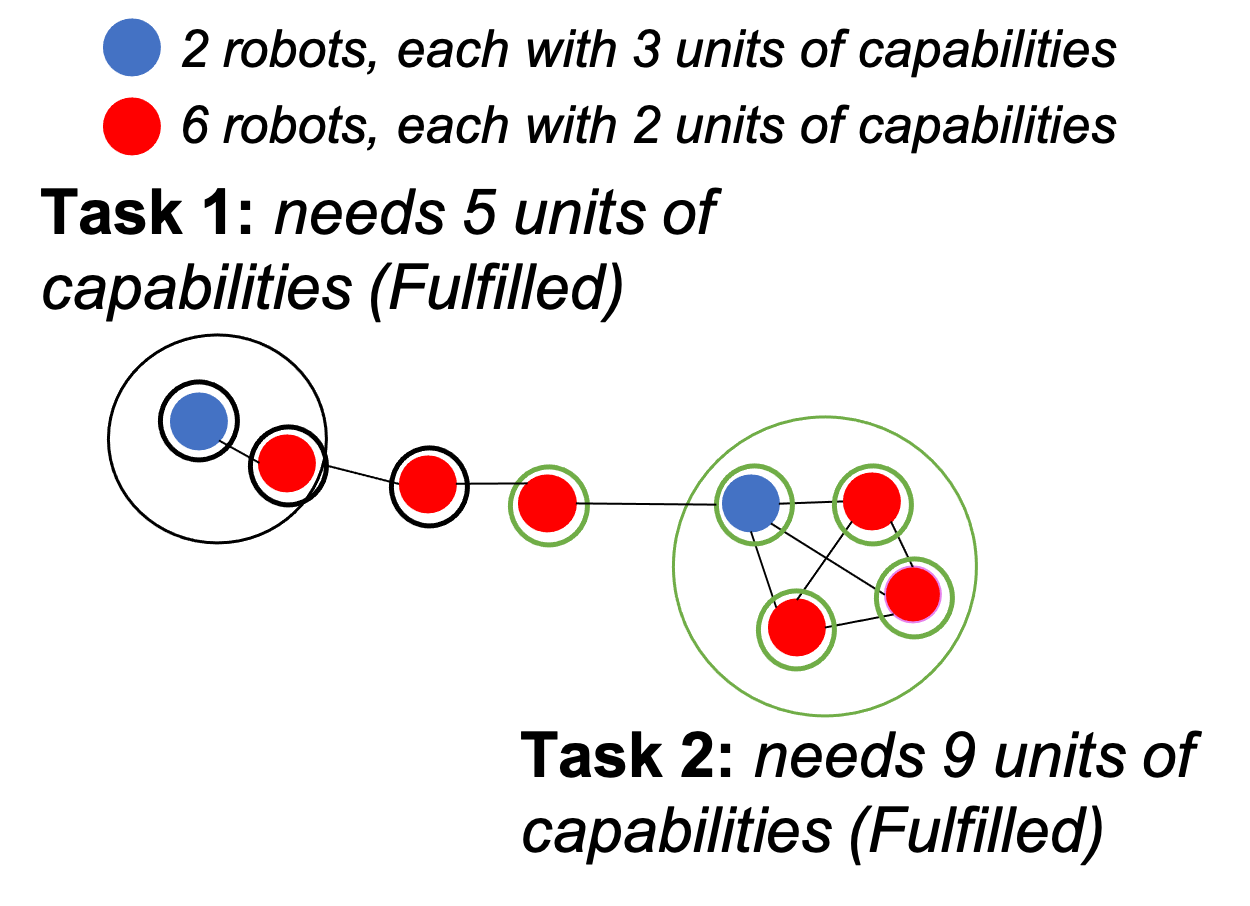}
\caption{Both tasks are fulfilled.}
\label{fig:introduction3}
\end{subfigure} \hfill
\caption{Illustration of issues brought by preserving connectivity within the robot team: 2 blue robots with 3 units of capabilities each and 6 red robots with 2 units of capabilities each are allocated to two tasks.  Task 1 requires 5 units of capabilities, and Task 2 needs 9 units. Both requirements should be overly fulfilled by 2 units. However, because of the connectivity constraints, in (a) Task 1 is under-allocated by 2 units, and  (b) Task 2 is under-allocated by 2 units. Only the configuration shown in (c) can satisfy both demands.}
\label{fig:introduction}
\end{figure*}

The heterogeneity of the multi-robot system in the work is quantified by different units of capabilities in various categories like sensor, mobility, payload, fuel capacity, etc.
Since the number of tasks and their requirements are not known a priori but are discovered as the robots move, the robots may need to be redistributed as new tasks are discovered. To achieve collaboration, robots are expected to exchange messages and share information, which requires the robot members to stay connected within at least one neighbor's communication range. Additionally, maintaining connectivity of the whole multi-robot system is crucial to ensure that the various robot groups that are assigned to different tasks can rejoin with one another after they have finished their tasks, so as to be re-allocated as new tasks arise. However, such connectivity constraints can impede the multi-robot system from following the planned allocation. Figure \ref{fig:introduction} is an illustration of the issue that arises from the connectivity constraints. In all three scenarios, the capabilities of one category provided by the allocated robots surpass the demands of the tasks. Nonetheless, some robots might never be able to reach their assigned target areas because of the connectivity constraints as shown in Fig. \ref{fig:introduction1} and Fig. \ref{fig:introduction2}. In those situations, we want the robots to be effectively redistributed to the configuration shown in Fig. \ref{fig:introduction3}. In this reallocation the traveling time of the candidate robots that are to be re-assigned to the new target must also be considered.

Most of the existing works deal with static task allocation or dynamic assignment at an abstract level \cite{Prorok:2016:FastRedistribution}, \cite{liu:2019:submodular}, where the detailed configuration of the robot team is usually not addressed.
The \textbf{contributions} of our work are: (1) it presents a systematic framework that combines combinatorial task allocation with the low-level control to generate different robot configurations with connectivity constraints that maximizes the overall system utility;
(2) A problem with a dynamic environment is proposed and solved by applying our framework; (3) The sub-optimality proof of the greedy assignment is provided, along with the simulation results and the numerical analysis.

\section{RELATED WORK}
\label{sec:related_work}

\indent For multi-robot systems, task allocation is a crucial and extensively studied topic \cite{Brian:2004:task_allocation_taxonomy}, \cite{Korsah:2013:task_allocation_taxonomy2}. In a Multi-Robot Task Allocation (MRTA) problem \cite{Korsah:2013:task_allocation_taxonomy2}, multiple tasks with different priorities require various capabilities from heterogeneous robots 
\cite{Zlot-2006-9632}. 
The problem is formulated as an NP-hard combinatorial optimization problem \cite{korte:2012:combinatorial}. Such problems can commonly be solved by mixed integer nonlinear programming (MINLP) \cite{lee2011mixed}, or be formulated into a distributed constraint optimization problems (DCOPs) \cite{Yoko:1991:DCOP} that can be solved by self-adjusting algorithms like MGM \cite{pearce:2007:MGM} and the Distributed Stochastic Algorithm (DSA) \cite{ZHANG:2005:DSA}, \cite{Lisy:2010:DCOP_mobile_sensing}. These distributed methods can solve the problem rapidly with good scalability. Auction-based algorithms are also widely applied. For example, combinatorial auction algorithms are used in allocation of virtual machine instances \cite{Duan:2017:crowdsensing}, and sequential single-item auction has been deployed on multi-robot systems \cite{koenig2006power}. We further develop the adaptive greedy algorithm based on the auction methods to better fit in our heterogeneous multi-robot dynamic redistribution problem.

Along with the algorithms that solve the task allocation problems for heterogeneous groups, many papers present methods on the dynamic redistribution to adapt the failures of the robots or the modifications of the knowledge about tasks in a changing environment. In \cite{notomista:2019:optimal}, the authors utilized constraint-based optimization to dynamically allocate heterogeneous robots along with the task execution controls. In \cite{Prorok:2016:FastRedistribution} and \cite{Prorok:2016:Formalizing}, the reallocation is characterized as transition rates for the robot groups based on their trait distribution. \cite{ramach:2019:resilience} presents a method to quickly reconfigure the network of the heterogeneous multi-robot system when their shared resources fail. In \cite{halasz:2007:dynamic}, the authors presented an approach to dynamically reassign the homogeneous robots to multiple locations in accordance with the changes in the environment. However, these dynamic redistribution strategies often neglect the actual arrangement and controls of the robot team due to the existence of connectivity and collision avoidance constraints. In contrast, our paper studies the case when multiple robots are executing different tasks with diverse importance and requirements and have connectivity constraints. Both the allocation plan and control execution are produced adaptively to meet the combinatorial task requirements.

\section{PROBLEM DEFINITION}
\label{sec:prob}

Consider a multi-robot system with $n$ heterogeneous robots that operate in a planar space. Denote the robot set as $\mathcal{A} = \{1,2,...,n\}$. The heterogeneity of each robot $i$ is quantified by its capability $\mathbf{c}_{i} \in\mathbb{R}^{o}$, and $o$ is the number of capability categories considered in the mission. The location of each robot is defined by a vector $\mathbf{x}_i\in\mathbb{R}^2$. In the problem, $m$ tasks are given and the task set is encoded as $\mathcal{J}= \{1,2,...,m\}$ 
\footnote{Since we solve the assignment problem at each time step, the number $m$ of tasks that are present is known but changing as execution proceeds. }. The location of each task $j$ is denoted as $\mathbf{y}_j\in\mathbb{R}^2$, which is dynamic throughout the process. The task assignment is defined as a set $\mathcal{S}$ containing robot-task pairs. It is equivalent to a binary matrix $S\in\mathbb{R}^{n\times m}$, and robot $i$ is assigned to task $j$ if and only if $S_{ij} = 1$. Since each robot can only be allocated to at most one task, for all robots, the number of tasks that the robot is allocated to is not more than one. Naturally, we denote the set of robots assigned to task $j$ as $\mathcal{A}_j$. Notice that, at the beginning the robots are all assigned to task 0 to indicate that they are available. The importance of each task $j$ is evaluated as a non-negative constant $v_j$, and the numbers of its required capability units in various categories are given as a vector $\mathbf{w}_{j}\in\mathbb{R}^{o}$. At each time step, we assess the remaining required capability of task $j$ as $\mathbf{r}_{j}\in\mathbb{R}^{o}$, whose entries are always larger than or equal to $0$. The robots are expected to execute the task the moment they enter within range $L$ from the task. 
Notice that $L$ can be arbitrarily small if the execution is actuated at the exact task location.

Our objective is to achieve an adaptive distribution so that the overall effective utility of the robot team can be maximized and allocation to the newly discovered targets can be solved while, in addition to the required capabilities and target importance, we also consider traveling costs and connectivity constraints.
Before the exploration, the importance of tasks along with their requirements are unknown. \textit{Notice that by exploration we mean one of the robots with sensing capability reaches the unknown task area with known position. We are not doing traditional exploration that maximizes the area coverage as it is out of the scope of this paper.} The importance of an unexplored task is a user-defined parameter deciding the priority of exploring new tasks comparing to staying and executing the current tasks.
As robots start to move, they discover new targets. Target discovery requires at least one robot with sensing ability to visit the target $j$ to acquire the information on the targets importance $v_j$ and required capabilities  $\mathbf{w}_j$. At every time step,
the remaining requirement of task $j$, $\mathbf{r}_j$, is calculated as the original requirement $\mathbf{w}_j$ minus the sum of capabilities of all the robots that are executing the task and is lower bounded by 0. The discovered target's position is given as a 2-dimensional vector $\mathbf{y}_j$.
\textit{Notice that the number of targets $m$ is known at every time step when solving the allocation problem,  but it is not a fixed constant for the whole problem horizon because new tasks can appear in the middle of mission executions.}
The goal is to minimize the weighted sum of the total remaining requirements along with the
traveling distance of each robot. To achieve that, an optimal assignment matrix $S^*$ is produced and controls $\mathbf{u}^* \in \mathbb{R}^{2n}$ are provided to distribute the robots. To simplify the dynamics model, in this work, $\mathbf{u}^* \in \mathbb{R}^{2n}$ is presented as the joint control input directly on the velocity. The objective is expressed as 

{\footnotesize
\begin{align}
    S^*,\mathbf{u}^* = &\argmin_{S,\mathbf{u}} \sum_{j = 1}^m (v_j\sum_{t=1}^o \mathbf{r}_{jt} - \alpha \sum_{i \in \mathcal{A}_j} h_{ij})\\\nonumber
      =&\argmin_{S,\mathbf{u}} \left(J_r - J_c\right)\\\nonumber
 =&\argmax_{S,\mathbf{u}}   \sum_{j = 1}^m (v_j\sum_{t = 1}^o(\min(\mathbf{w}_{jt}, \sum_{i\in \mathcal{A}_j} \mathbf{c}_{it}))  + \alpha \sum_{i \in \mathcal{A}_j} h_{ij})\\\nonumber
  =&\argmax_{S,\mathbf{u}} \left(J_e +  J_c\right)
\end{align}
}where $h_{ij}$ is defined as $h_{ij}=\frac{1}{1+\Vert\mathbf{x}_i - \mathbf{y}_j \Vert_2}$.

In the objective function, $J_r = \sum_{j = 1}^m (v_j\sum_{t=1}^o \mathbf{r}_{jt})$ is the remaining utility of the system, which is to be minimized, $J_e = \sum_{j = 1}^m (v_j\sum_{t = 1}^o(\min(\mathbf{w}_{jt}, \sum_{i\in \mathcal{A}_j} \mathbf{c}_{it})))$ is the effective utilities of the robots, which is to be maximized, and $J_c = \alpha\sum_{j = 1}^m  \sum_{i \in \mathcal{A}_j} h_{ij}$ is the inverse of the traveling costs with normalization, in which $\alpha$ is a scaling constant that determines how important the traveling cost is in our problem.  The traveling costs are formulated this way so that there is a sub-optimality bound for our algorithm.

\section{METHODS}
\label{sec:methods}
In this section, our framework to obtain an sub-optimal allocation plan and controls for the problem defined in Section \ref{sec:prob} is described as following: In Section \ref{sec:auction}, we design a greedy algorithm that can quickly and adaptively generate a sub-optimal assignment at each time step, which is passed to the control-level optimization for further improvement. 
By proving the submodularity and monotonicity of our objective function in Section \ref{sec:analysis}, we can deploy a greedy algorithm that has known suboptimality bound.
In Section \ref{sec:barrierCertificate}, we review a connectivity maintenance technique using barrier certificate \cite{BORRMANN:2015:BarrierCertificates}, which is utilized as the constraints of our control optimization. In Section \ref{sec:algorithm}, we present a control optimization framework, which extends the Behavior Mixing scheme, to achieve the desired configurations. 

\subsection{Greedy Dynamic Task Allocation } 
\label{sec:auction}
At every time step, we want to first obtain a allocation plan served as a baseline for the control rendering. Since task allocation is a strongly NP-hard problem \cite{korte:2012:combinatorial}, greedy methods are efficient and flexible choices with provable optimality \cite{Fisher:1978:submodular, Ryan:2017:matroid, Shin:2019:sample_greedy} and polynomial computational complexity. In our work, the adaptive greedy algorithm is constructed to provide both the initial assignment and the adaptive re-assignments as following:

\begin{algorithm}\footnotesize
\SetAlgoLined
\KwResult{The assignment matrix $S$}
\While {$\mathcal{A}$ is not empty}{
 \For{robot $i \in \mathcal{A}$ }{
    \For{task $j$ = 0 : $m$}{
        $g_{ij} = v_j\sum_{t = 1}^o(\min(\mathbf{c}_{it},\max(0,\mathbf{w}_{jt})))$\;
        \eIf{$j'=0$}
        { $l_{ij} = 0$\;}
        {$l_{ij} = v_{j'}\sum_{t = 1}^o(\max(0,\min(\mathbf{w}_{j't} + \mathbf{c}_{it},\mathbf{c}_{it})))$\;}
        $h_{ij} = \frac{1}{1+\Vert \mathbf{x}_i - \mathbf{y}_j\Vert_2}$\;
        $p_{ij} = g_{ij} - l_{ij} + \alpha h_{ij}$\;
        }
        Find the maximal price $p_i$ from task $j_{\text{best}}$ to robot $i$, also known as the marginal gain\;
}

Collecting the reward $p_i$ from all the robots\;

Choose robot $i^*$ with the largest reward among all\; 
    \eIf{$p_{i^*} == \alpha$}
    {
     {break \;}
    }
   { $j^* = j_{\text{best}}$ \;}
      
      $S_{i^*j'} = 0$\;
   $S_{i^*j^*} = 1$\;
    $\mathbf{w}_{j^*} = \mathbf{w}_{j^*} - \mathbf{c}_{i^*}$\; 
    $\mathbf{w}_{j'} = \mathbf{w}_{j'} + \mathbf{c}_{i^*}$\;
    $\mathcal{A} \setminus i^*$ \;
    $\mathbf{w}_0 = \mathbf{0}$\;
}
\caption{Adaptive Greedy Algorithm}
\label{algo:auction}
\end{algorithm}

From Line 2 to Line 12, the rewards of all possible robot-task pairs are determined.
The reward of matching task $j$ to robot $i$ is calculated as the summation of three terms gain $g_{ij}$, loss $l_{ij}$, and cost $h_{ij}$ (Line 4 - 10), where $j'$ is the original task that robot $i$ was assigned to. If robot $i$ is available, $j' = 0$ and its loss term is 0. Then, the reward $p_{ij}$ is calculated as $p_{ij} = g_{ij} - l_{ij} + \alpha h_{ij}$,
where $\alpha$ is a scaling constant as stated in the explanations of the objective function.

In Line 13-16, the maximal reward among all pairs is found. Notice that the lower bound of $p_i$ is $\alpha$. If the maximal reward is equal to $\alpha$, then the best strategy for all the remaining robots in set $\mathcal{A}$ is to stay with their original tasks and the assigning process can be early terminated (Line 18).
A robot will be assigned to a task out of its previous assignment only when the improvement in overall utility is greater than the traveling cost (Line 20). After that, the assignment matrix $S$ is updated, and so is the requirement vector $\mathbf{w}$. The selected robot $i^*$ is removed from the robot set $\mathcal{A}$, and the demand of task 0 (stay still) is reset back to 0 (Line 22-27). The assignment process takes at most $n$ rounds and for each round it requires $mn$ reward evaluations. Thus, it can be easily proven that the computational complexity of the algorithm is $O(mn^2)$ in the worst case. However, in most scenarios since the algorithm is early terminated, the complexity is $O(1)$ when there is no update about tasks' information.

\subsection{Algorithm Performance Analysis}
\label{sec:analysis}
In this section, we show that using this greedy algorithm to maximize the objective function, the solution quality is bounded by proving the submodularity and the monotonicity of our objective function. Let $\mathcal{V}$ be a finite set, which is also the grounded set of our problem. Denote the optimal assignment as $\mathcal{S}_{\text{opt}} \subseteq \mathcal{V}$.  The objective function as defined before is $f(\mathcal{S}^*) = \sum_{j = 1}^m \left(v_j\sum_{t = 1}^o(\min(\mathbf{w}_{jt}, \sum_{i\in \mathcal{A}_j} \mathbf{c}_{it}))  + \alpha \sum_{i \in \mathcal{A}_j} h_{ij}\right)$, which is always non-negative ($f(\emptyset) = 0$).
\begin{corollary}
For the solution $\mathcal{S}^*$ obtained from our greedy algorithm, it is guaranteed that $f(\mathcal{S}^*) \geq \frac{1}{2}f(\mathcal{S}_{\text{opt}})$
\label{corollary:bound}
\end{corollary}

To prove this corollary, we can apply the conclusion from Fisher's work \cite{Fisher:1978:submodular} that if a function $f:2^{\mathcal{V}}\rightarrow \mathbb{R}$ is submodular and monotone, the greedy algorithm (taking the largest marginal gain at each step) is guaranteed a constant optimality bound as 1/2.
Submodularity and monotonicity are defined as below.

\begin{definition}(Submodularity \cite{Fujishige:2005:submodular})
\label{definition:submodular}
A function $f:2^\mathcal{V} \rightarrow \mathbb{R}$ is submodular if and only if for any $\mathcal{X} \subseteq \mathcal{Y} \subseteq \mathcal{V}$, $\forall e \in \mathcal{V}\setminus \mathcal{Y}$, the following inequality holds
\begin{align}
 \Delta_f(e|\mathcal{X}) \geq \Delta_f(e|\mathcal{Y}) 
 \label{eqn:submodular}
\end{align}
where $\Delta_f(e|\mathcal{S})$ is defined as 
\begin{align}
    \Delta_f(e|\mathcal{S}) = f(\mathcal{S} \cup \{e\}) - f(\mathcal{S})
    \label{eqn:margin}
\end{align}
\end{definition}

\begin{definition}(Monotonicity \cite{Fujishige:2005:submodular})
\label{definition:monotone}
A function $f:2^\mathcal{V} \rightarrow \mathbb{R}$ being monotonic is equivalent to show that 
\begin{align}\footnotesize
    \forall \mathcal{X}\subseteq \mathcal{Y}\subseteq \mathcal{V}, f(\mathcal{Y}) - f(\mathcal{X}) \geq 0
\end{align}
\end{definition}
Because of the page limit, the detailed proofs are not explicitly provided here.
With both properties proven, applying the conclusion from \cite{Fujishige:2005:submodular}, the optimality bound of this algorithm is proven to be $f(\mathcal{S}^*) \geq \frac{1}{2}f(\mathcal{S}_{\text{opt}})$ as stated in Corollary \ref{corollary:bound}.

\subsection{Connectivity Constraints using Barrier Certificates}
\label{sec:barrierCertificate}
To collaborate, a multi-robot system is required to stay connected so that the robots can send  peer to peer messages to each other \cite{luo2020behavior, luo2019minimum, luo2019voronoi}. The framework of Behavior Mixing \cite{luo2020behavior} is employed to enable the robots to split into subgroups while the global connectivity is preserved so that, after each subgroup has finished its task, it can rejoin the other groups. A particular communication sub-graph $\mathcal{G}^c\subseteq \mathcal{G}$ of the multi-robot system is maintained, where $\mathcal{G} = (\mathcal{V,E})$ is the current communication graph with each vertex $v \in \mathcal{V}$ as a robot and each undirected edge $(v_i,v_j) \in \mathcal{E}$ as the communication link if the distance between the two robots $i,j$ is equal or less than the limited communication range $R_c$. At each time step, based on the current configuration, a minimum connectivity constraint spanning tree (MCCST) $\mathcal{G}^c\subseteq \mathcal{G}$ is built \cite{luo2020behavior, luo2019voronoi} so that the constructed Minimum Spanning Tree denotes the connectivity topology to preserve that incur minimally disruptive connectivity constraints at the moment. With that, we can define the set of $\mathbf{x}$ for which the connectivity can always be preserved using the following expressions:
{
\begin{align}
    h_{i,j}^c(\mathbf{x}) &= R^2_c - \Vert \mathbf{x}_i - \mathbf{x}_j\Vert_2^2\\
    \mathcal{H}^c_{i,j} &= \{ \mathbf{x} \in \mathbb{R}^{2n} : h_{i,j}^c \geq 0\}
\end{align}
}With the MCCST graph $\mathcal{G}^c = (\mathcal{V},\mathcal{E}^c)$, we can create a feasible set for $\mathbf{x}$ so that any connected pairs in $\mathcal{G}^c$ should be in the set $\mathcal{H}^c(\mathcal{G}^c)$, which is expressed as
 {
 \begin{align}
     \mathcal{H}^c(\mathcal{G}^c) = \underset{(v_i,v_j)\in \mathcal{E}^c}{\cap} \mathcal{H}^c_{i,j}
 \end{align}
 }Following the barrier certificate function used for collision avoidance in \cite{BORRMANN:2015:BarrierCertificates, luo2020multi}, we can write a barrier function for connectivity maintenance as 

{
\footnotesize
\begin{align}
    \mathcal{B}^c(\mathbf{x},\mathcal{G}^c) = \{\mathbf{u} \in \mathbb{R}^{2n} : \dot{h}^c_{i,j}(\mathbf{x}) + \gamma h^c_{i,j}(\mathbf{x}) \geq0, \forall (v_i,v_j) \in \mathcal{E}^c\}
\end{align}
}where $\gamma$ is a user-defined parameter to enclose the available set, and $\mathbf{u}$ is the joint control input directly on the velocity. It is proven that $\mathcal{H}^c$ is a forward invariant set if $\mathbf{u}$ stays in $\mathcal{B}^c$ and are \emph{least constraining} on controllers to maintain global connectivity \cite{luo2020behavior, luo2019voronoi}. This equation will serve as the connectivity constraint in our optimization for controls, which is discussed in the next section.

\subsection{Redistribution with Heterogeneous Robots}
\label{sec:algorithm}

As we mentioned in Section \ref{sec:intro}, because of the connectivity constraints, even given a good allocation, the tasks still cannot be carried out as desired. For example, when a new task appears, one of robots will be assigned to explore it. Yet, constrained by all of its neighbors, it is impossible for this robot to examine the new task, unless there is a mechanism to help the robot drag its neighbors to the new goal while ensuring that the majority of the system is still performing the tasks. Here, we develop a method to achieve an adjustment in the distribution of the robots at the control level. 

Without the connectivity constraints, assuming all the robots can reach their assigned targets, we calculate the remaining requirement of each task $j$ in category $t$ as 
{
\begin{align}
    \mathbf{r}_{jt} = \max(0,\mathbf{w}_{jt} - \sum_{i\in A_j} \mathbf{c}_{it}) 
\label{eqn:ideal_remain}
\end{align}
}As mentioned in Section \ref{sec:prob}, here we assume the task is executed when robots rendezvous within execution range $L$ around the task. Also, as a result of the connectivity constraints, not all robots are able to arrive at their allocated areas. In other words, the true remaining requirements $\mathbf{\hat{r}}_{jt}$ can be different from the ideal remaining requirements $\mathbf{r}_{jt}$ shown in Eqn. \ref{eqn:ideal_remain}.
The true remaining requirement of task $j$ in category $t$ is calculated as 
\begin{align}
    \mathbf{\hat{r}}_{jt} =\max(0,\mathbf{w}_{jt} - \sum_{i\in A_j}\mathbf{c}_{it} H (L-d_{ij}) ) 
\end{align}
where $d_{ij}$ is the $l^2-$ norm of the distance between the current position of robot $i$ and the position of its assigned task $j$, 
and $H$ is a Heaviside Step Function \cite{abramowitz:1988:handbook} that outputs 1 if the input is non-negative and 0 otherwise.

With that in mind, we calculate controls by employing the quadratic programming (QP) as following,

{\footnotesize
\begin{align}
    \label{eqn:qp}
     \mathbf{u}^* = \argmin_{\mathbf{u}} \sum_{i=1}^n (a_i+1)\Vert \mathbf{u}_i-\hat{\mathbf{u}}_i \Vert^2\quad 
    \text{s.t.} \quad  \mathbf{u} \in \mathcal{B}^c(\mathbf{x},\mathcal{G}^c)  
\end{align}
}where $\hat{\mathbf{u}}_i$ is the primary controller for robot $i$'s assigned task and $\mathbf{u}^* \in \mathbb{R}^{2n}$ contains the control inputs of all robots. For simplicity without losing generality, single integrator dynamics is used for the task-related controller of each robot, i.e., $\hat{\mathbf{u}}_i = -K_p(\mathbf{x}_i-\mathbf{y}_j)$. A constant ``$1$" is added to the coefficient $a_i$ to avoid the lack of ranks in quadratic programming when $a_i = 0$.

Here, we want to to maximize the utility function, which is equivalent to reducing the weighted remaining unfulfilled requirements
by assigning higher weights to the robots whose assigned tasks are less fulfilled and of higher priority. For each robot $i$ that is assigned to task $j$, its coefficient $a_i$ in the QP is calculated as
{
\begin{align}\label{eq:ai_sim}\
    a_i = v_j \sum_{s=1}^o \mathbf{c}_{it} \mathbf{\hat{r}}_{jt}
\end{align}
}Thus, how much the robot's primary controller is preserved depends on the value and the remaining unfulfilled demands of its assigned task, and the capability of the robot, which indicates how much improvement the robot can contribute to the system. The more capable robot will be encouraged to fulfill the gap of the more important and less fulfilled task, and others will tend to serve as connectivity nodes.

When solving the QP, it is important that the objective function is differentiable. Now we have a step function $H$ in the coefficient, and thus the robots might bounce back and forth near the boundaries. To avoid that, we want to replace it to be a differentiable function. A scaled and shifted sigmoid function $\sigma$ is used here to replace the Heaviside step function. So we can re-write (\ref{eq:ai_sim}) as follows.
{
\begin{align}
        a_i 
        = v_j \sum_{s=1}^o \mathbf{c}_{it}[\max (0,\mathbf{w}_{jt} - \sum_{i\in A_j} \mathbf{c}_{it} \sigma (k(L-d_i)+b) )]
\end{align}  
}where $k$ is a scale factor, and $b$ is a shift constant.

\section{RESULTS}
\label{sec:res}

\subsection{Simulation Result}
\label{sec:sim_res}
\begin{figure*}[t]
\centering
\begin{subfigure}[t]{0.23\textwidth}
\includegraphics[width=\textwidth]{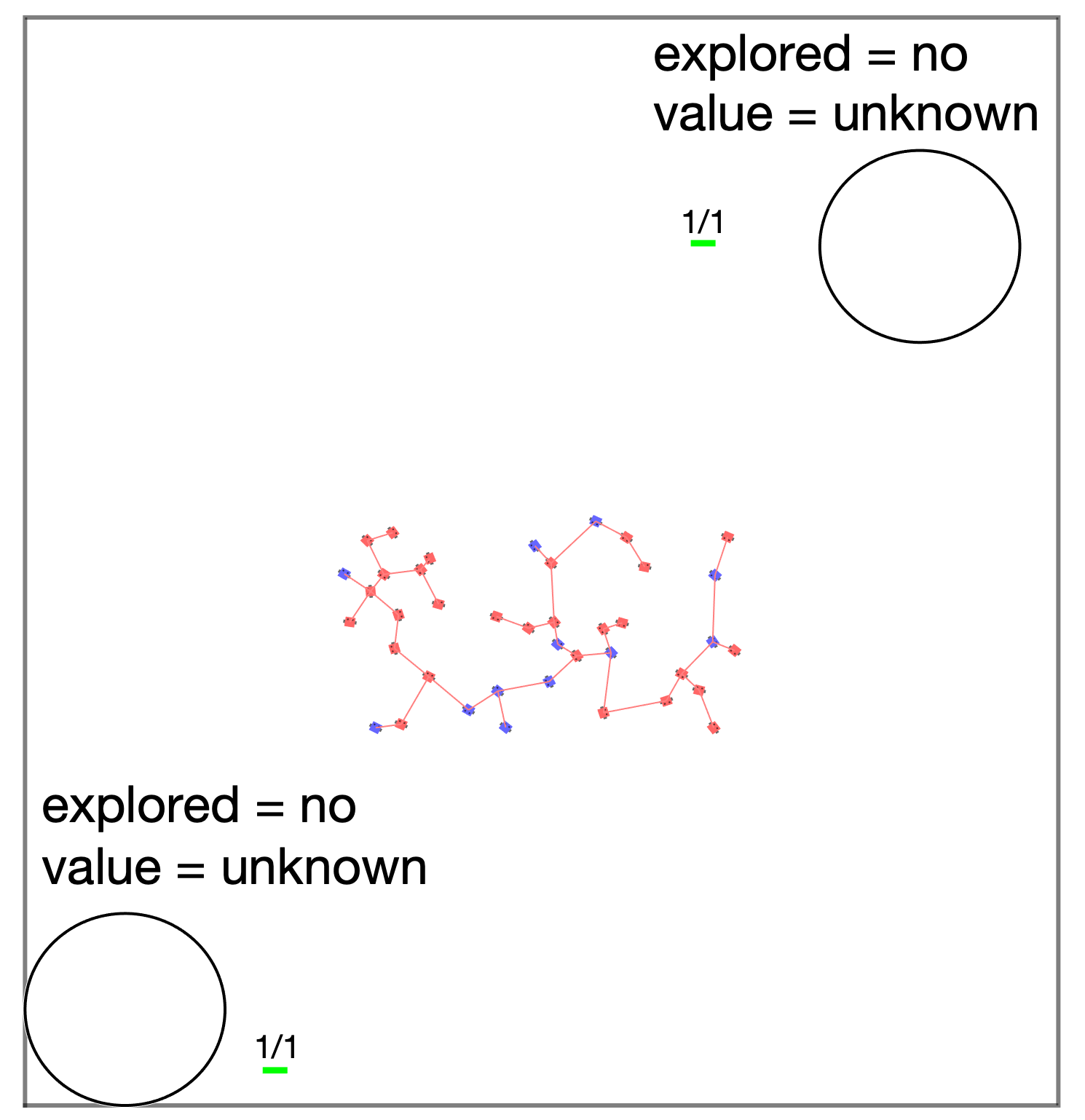}
\caption{Time Step = 0 (Only two tasks are present. Each task requires only one unit of green capability to fully acquire the information of it)}
\label{fig:initial}
\end{subfigure} \hfill
\begin{subfigure}[t]{0.23\textwidth}
\includegraphics[width=\textwidth]{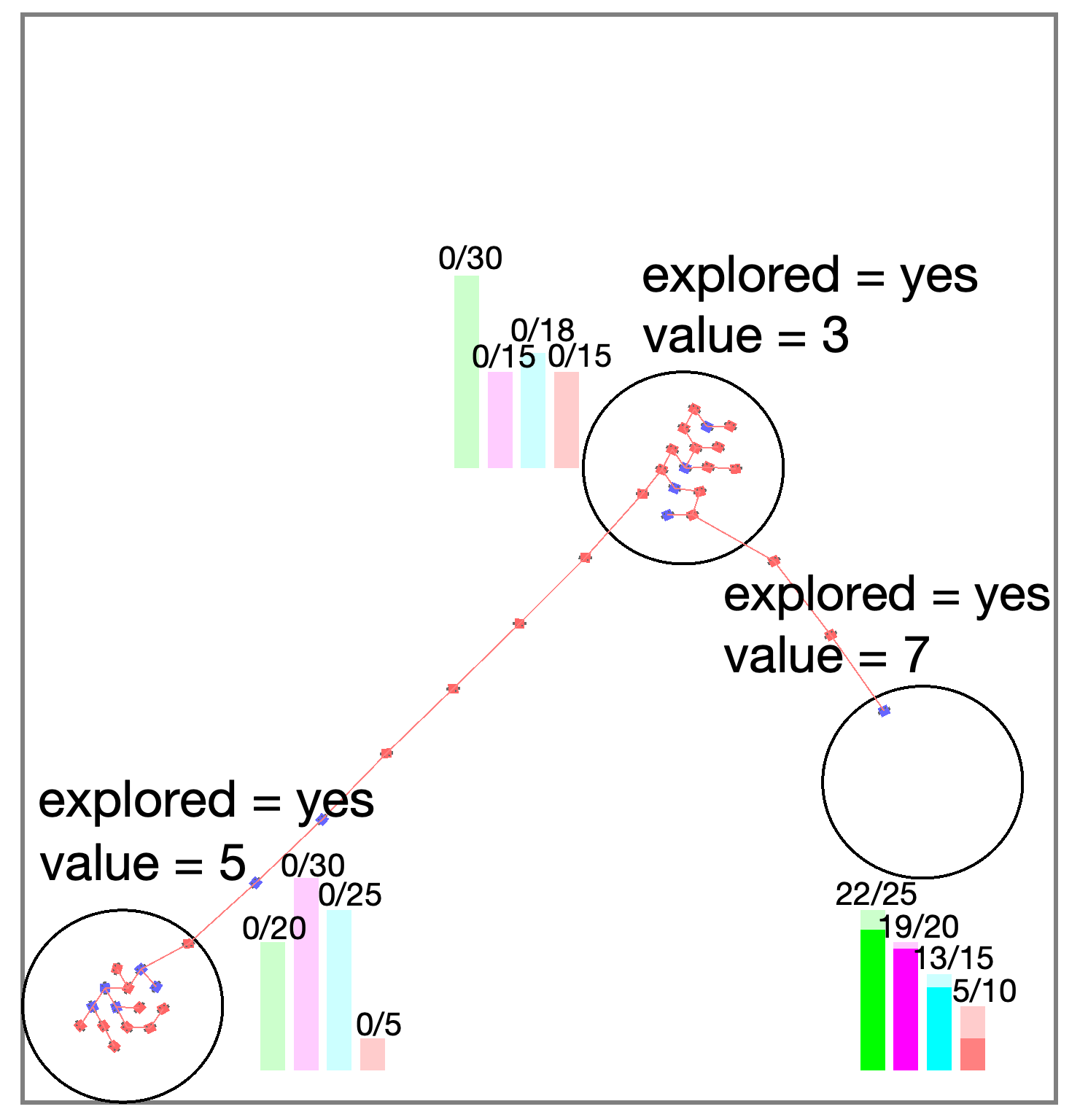}
\caption{Time Step = 1620 (The newly appeared task is explored and its value and requirements are exposed to the system)}
\label{fig:res0}
\end{subfigure} \hfill
\begin{subfigure}[t]{0.23\textwidth}
\includegraphics[width=\textwidth]{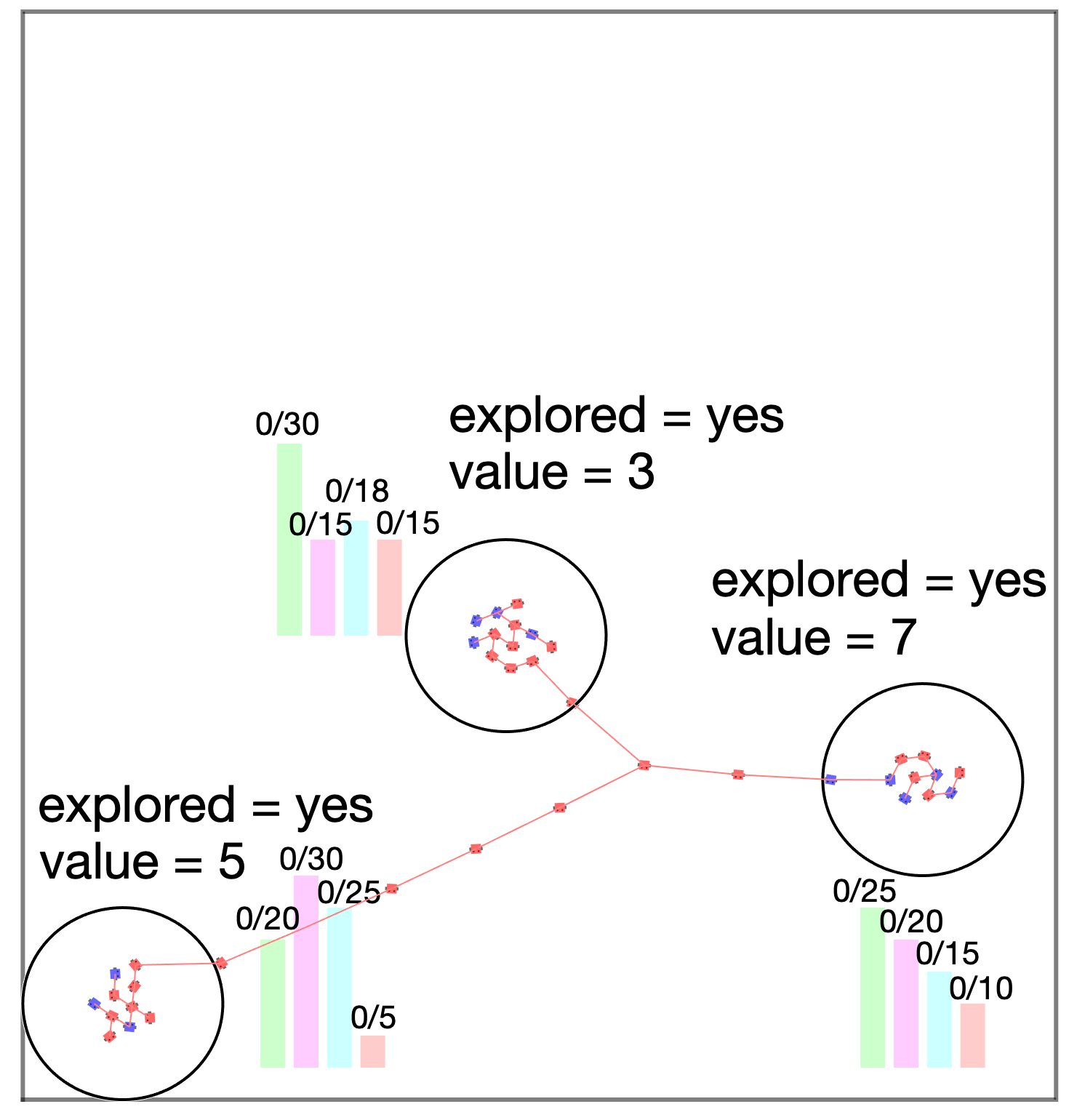}
\caption{Time Step = 3270 (Final configuration when the three tasks are close to each other)}
\label{fig:res1}
\end{subfigure} \hfill
\begin{subfigure}[t]{0.23\textwidth}
\includegraphics[width=\textwidth]{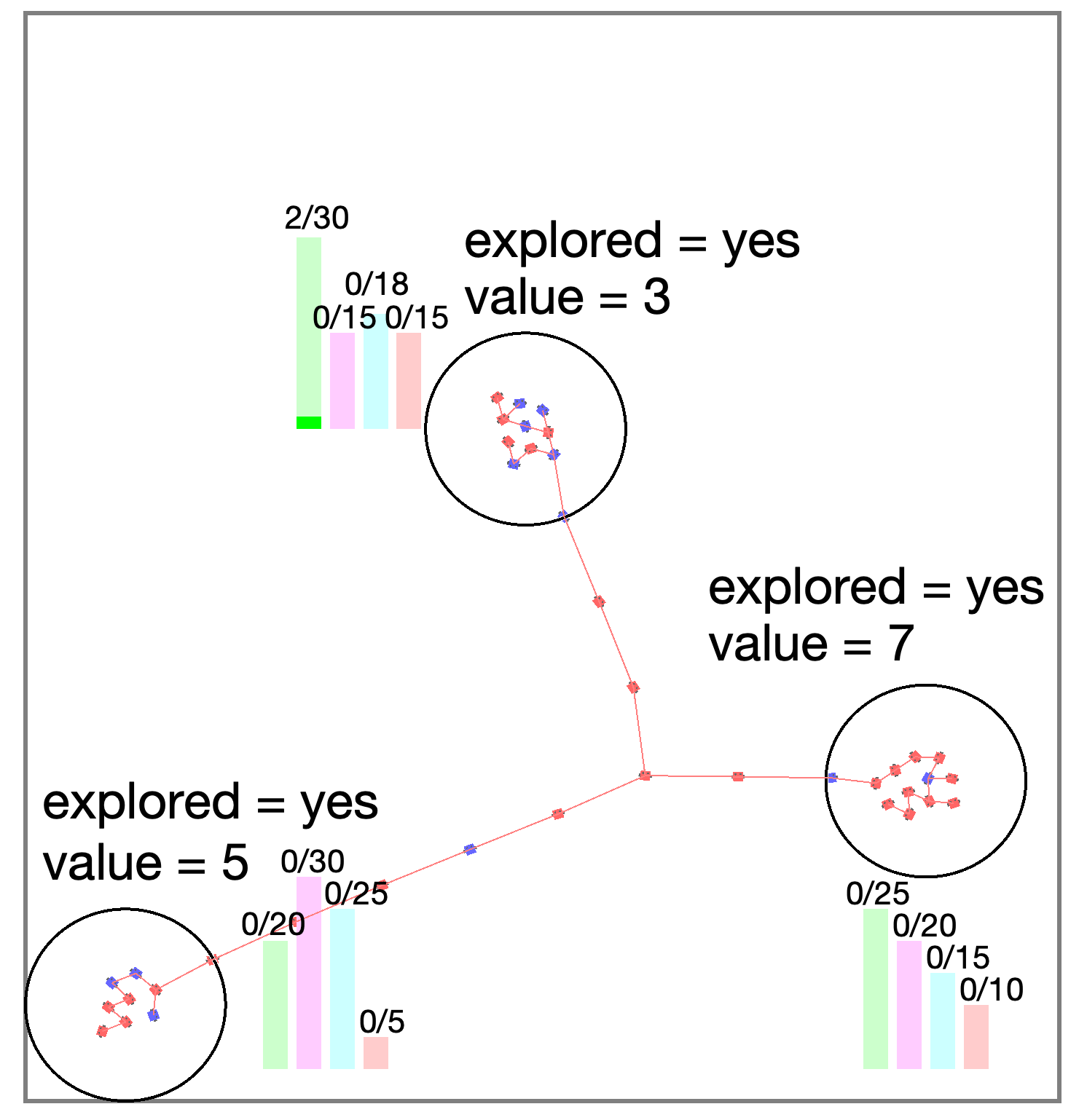}
\caption{Time Step = 5500 (Final configuration when the three tasks are farther away from each other)}
\label{fig:res2}
\end{subfigure} \hfill
\caption{Four stages of the simulation test running on MATLAB. Two types of robots, three tasks, and four types of capabilities are present in the test. }
\label{fig:result}
\end{figure*}
One simulation test case is demonstrated in Fig. \ref{fig:result}, which is also presented more thoroughly in the attached video. The importance and the exploration state is displayed on the top of each state.
Before explored, the importance $v$ of the unknown tasks are chosen to be $50$, which is a big incentive to drive the system to explore them. The requirement of each unknown task is one unit of the green capability, and the weight on traveling costs $\alpha = 1.0$. We have 12 blue robots and 28 red robots in total. Each blue robot owns 3 units of the green capability, 1 unit of the magenta capability, 2 units of the cyan capability, and 5 units of the orange capability. Each red robot owns 2 units of the green capability, 5 unit of the magenta capability, 4 units of the cyan capability, and no orange capability. At the beginning, two tasks are present with unknown importance $v$ and requirements $\mathbf{w}$ initially and must be explored. As shown in Fig. \ref{fig:initial}, before exploration, each task only requires one unit of green capability, e.g., the sensing ability, to fully obtain the knowledge of the task, i.e., the importance and the requirement. The top right task will move around and the system needs to track it and reconfigure the deployment to maximize the overall effective utility. During the mission execution, a third task appears at bottom right so the system needs to adapt to the change dynamically. Figure \ref{fig:res0} demonstrates the state when the task at the lower right corner is just explored. The solid color bars next to the tasks indicate the remaining requirements $\mathbf{r}$ of different types of capabilities, and the transparent color bars indicate the original requirements. The fraction number on the top of each color bar represents the remaining required units of capabilities out of the required ones. The goal is to maximize the overall effective utility $J_e$ and minimize the remaining utility $J_r$, thus, bringing down the solid color bars. 

Figure \ref{fig:res1} shows the configuration of the multi-robot system when the three targets are close together. Therefore, less robots are needed to serve as the connectivity nodes are more robots can actually contribute to the tasks. In this case, all the demands are satisfied, the remaining utility is $0$, and thus the overall effective utility is maximized. 
When the task in the middle moves away from the other two tasks, the number of robots is not large enough to cover all the requirements from the tasks plus satisfying the connectivity constraints. Figure \ref{fig:res2} is to demonstrate that in such scenarios, this framework will drive the multi-robot system to the configuration that minimizes the unfulfilled requirements weighted by their importance. We can observe from the figure that, in total 8 robots have to stay in the open space to reserve the connectivity. Because of the dynamic redistribution algorithm, the system is reconfigured so that only the task on the top, which is the least important task among the three, lacks 2 units of the green capability. The robots are allocated and distributed this way to minimize the remaining requirements weighted by tasks' importance, and thus maximizing the overall utility of the multi-robot system.

\subsection{Numerical Result}

To demonstrate the efficiency and the performance of our adaptive greedy allocation algorithm, we compare it with MIDACO \cite{Schlueter:MIDACO}, which is a commercial MINLP solver based on Ant Colony Optimization algorithm \cite{Schlueter:ACO}, and the GA MINLP solver in MATLAB Toolbox \cite{MatlabOTB}.\footnote{The configuration parameters are as following: MaxStallGenerations $=50$, FunctionTolerance $=1e-10$, MaxGenerations $=3000$, others by default.} The numerical results are summarized in the following two figures. Notice that because of the limitations of other MINLP algorithms, the traveling distance is not taken into considerations here.

\begin{figure}[h]
\centering
\begin{subfigure}{0.24\textwidth}
\includegraphics[width=\textwidth]{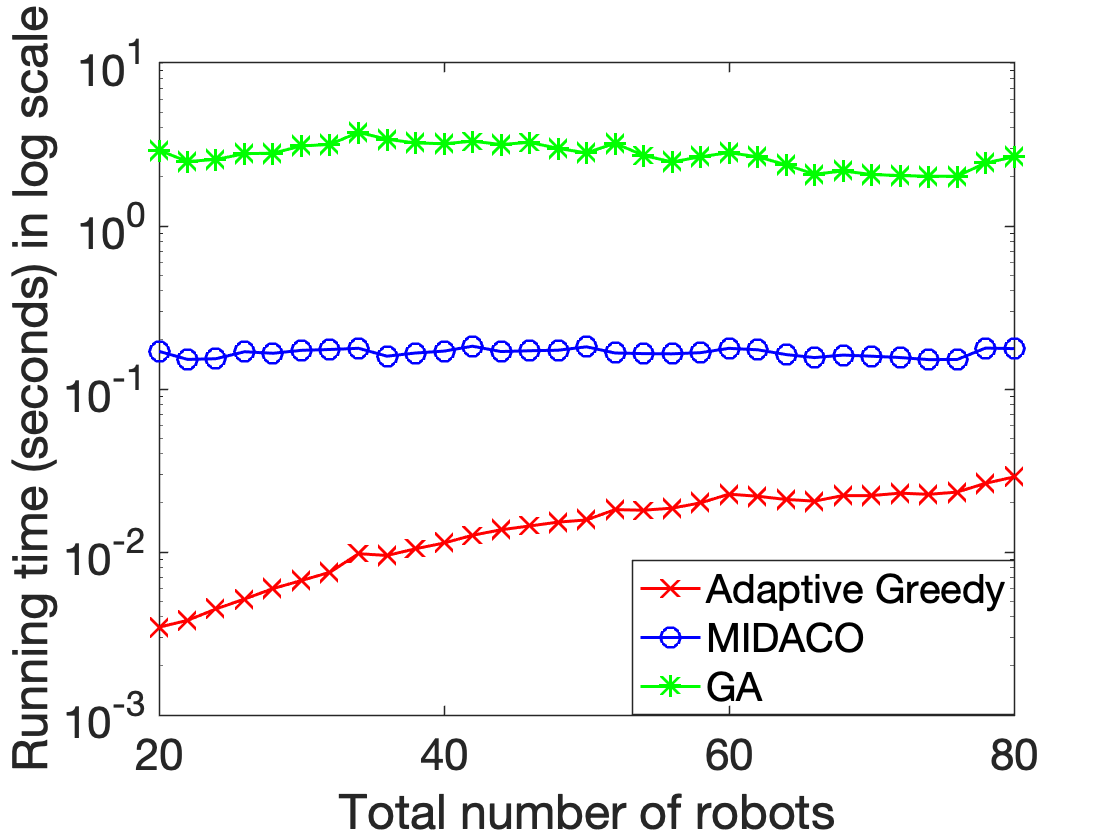}
\caption{Running time comparison in log scale}
\label{fig:num_res1}
\end{subfigure}
\begin{subfigure}{0.24\textwidth}
\includegraphics[width=\textwidth]{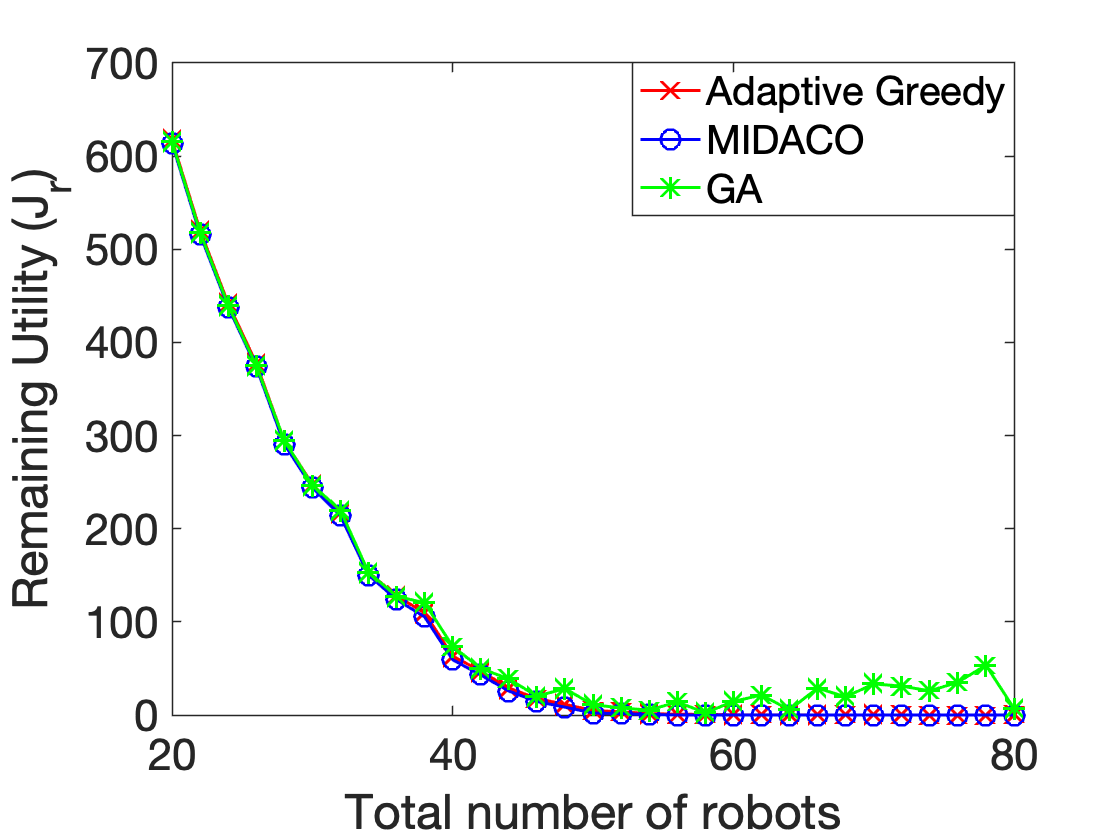}
\caption{Remaining utility $J_r$ comparison}
\label{fig:num_res2}
\end{subfigure} \hfill
\caption{Numerical results summary. In both figures, each data point represents the mean values of the 200 experiments.}
\end{figure}

In this comparison, we have the same setup as the simulation tests in terms of the types of robots, their capabilities, and the types of capabilities (green, magenta, cyan, orange). The total number of robots is the variant that increases from 20 to 80, half of which are the blue type robots and the other half are the red type. Three tasks are present in all the tests. For each data point in the plots, we generate 200 test cases and take the average value of them. In each test case, the requirement of each task in every category is randomly generated between 10 and 50, and the importance of every task is a random integer from 1 to 10.

We can observe from Fig.\ref{fig:num_res1} that regarding to the computational efficiency, our greedy method is tremendously faster than the other two. Because the adaptive greedy method solves the assignment problem for each individual robot, its running time increases quadratically with the number of robots. Still, it is at least 10 times faster than MIDACO and about 300 times faster than GA. In terms of the performance, which is measured by the remaining utility $J_r$, the greedy method is only slightly worse than MIDACO, while consistently better than GA. This gives our method a great potential to scale up the problem. It is also worth noticing that besides the scalability, our method has the advantage in flexibility. The traveling costs, which is a crucial factor in practice, can be easily incorporated into our method. Moreover, the assignments in our method are modified incrementally so the whole configuration of the multi-robot system will not be redesigned from scratch to adapt to the changes in the environment and most of the calculations will be early terminated as explained in Section \ref{sec:auction}. The advantages of our greedy method could be more significant when the types of robots and the number of tasks are scaled up as the computational cost of the greedy method does not depend on the number of types and only increases linearly with the number of tasks.\footnote{The free educational version of MIDACO only supports at most four design variables, so we did not extend the numerical tests further.}

\section{CONCLUSIONS}
\label{sec:conclusions}
In this paper, we present a task allocation method that dynamically distributes robots with different units of capabilities in various categories in a changing environment where tasks appear dynamically, while considering connectivity maintenance and the heterogeneity of controllers. We proposed an adaptive greedy algorithm that takes traveling costs into considerations with provable optimality bound and a redistribution mechanism at the control level to mitigate the loss due to the connectivity constraints. We show experimental results to demonstrate the efficiency of our algorithm in maximizing the utility of the robot team and the comparison of numerical performance between our method and the two commercial MINLP solvers.
Future work will include scheduling by adding the temporal dimension in task allocation to further reduce the costs resulted from change of targets during traveling. Also, we expect to deploy this framework on a physical robot system.

\addtolength{\textheight}{-7cm}   









\pagebreak

\bibliography{bibliography.bib}

\begin{thebibliography}{10}
\providecommand{\url}[1]{#1}
\csname url@samestyle\endcsname
\providecommand{\newblock}{\relax}
\providecommand{\bibinfo}[2]{#2}
\providecommand{\BIBentrySTDinterwordspacing}{\spaceskip=0pt\relax}
\providecommand{\BIBentryALTinterwordstretchfactor}{4}
\providecommand{\BIBentryALTinterwordspacing}{\spaceskip=\fontdimen2\font plus
\BIBentryALTinterwordstretchfactor\fontdimen3\font minus
  \fontdimen4\font\relax}
\providecommand{\BIBforeignlanguage}[2]{{%
\expandafter\ifx\csname l@#1\endcsname\relax
\typeout{** WARNING: IEEEtran.bst: No hyphenation pattern has been}%
\typeout{** loaded for the language `#1'. Using the pattern for}%
\typeout{** the default language instead.}%
\else
\language=\csname l@#1\endcsname
\fi
#2}}
\providecommand{\BIBdecl}{\relax}
\BIBdecl

\bibitem{Prorok:2016:FastRedistribution}
A.~Prorok, M.~A. Hsieh, and V.~Kumar, ``Fast redistribution of a swarm of
  heterogeneous robots,'' ser. BICT'15.\hskip 1em plus 0.5em minus 0.4em\relax
  ICST, Brussels, Belgium, Belgium: ICST (Institute for Computer Sciences,
  Social-Informatics and Telecommunications Engineering), 2016, pp. 249--255.

\bibitem{liu:2019:submodular}
J.~Liu and R.~K. Williams, ``Submodular optimization for coupled task
  allocation and intermittent deployment problems,'' \emph{IEEE Robotics and
  Automation Letters}, vol.~4, no.~4, pp. 3169--3176, 2019.

\bibitem{Brian:2004:task_allocation_taxonomy}
B.~P. Gerkey and M.~J. Matarić, ``A formal analysis and taxonomy of task
  allocation in multi-robot systems,'' \emph{The International Journal of
  Robotics Research}, vol.~23, no.~9, pp. 939--954, 2004.

\bibitem{Korsah:2013:task_allocation_taxonomy2}
G.~A. Korsah, A.~Stentz, and M.~B. Dias, ``A comprehensive taxonomy for
  multi-robot task allocation,'' \emph{The International Journal of Robotics
  Research}, vol.~32, no.~12, pp. 1495--1512, 2013.

\bibitem{Zlot-2006-9632}
R.~M. Zlot, ``An auction-based approach to complex task allocation for
  multirobot teams,'' Ph.D. dissertation, Carnegie Mellon University,
  Pittsburgh, PA, December 2006.

\bibitem{korte:2012:combinatorial}
B.~Korte, J.~Vygen, B.~Korte, and J.~Vygen, \emph{Combinatorial
  optimization}.\hskip 1em plus 0.5em minus 0.4em\relax Springer, 2012, vol.~2.

\bibitem{lee2011mixed}
J.~Lee and S.~Leyffer, \emph{Mixed integer nonlinear programming}.\hskip 1em
  plus 0.5em minus 0.4em\relax Springer Science \& Business Media, 2011, vol.
  154.

\bibitem{Yoko:1991:DCOP}
M.~Yokoo and E.~H. Durfee, ``Distributed constraint optimization as a formal
  model of partially adversarial cooperation,'' Tech. Rep., 1991.

\bibitem{pearce:2007:MGM}
J.~P. Pearce and M.~Tambe, ``Quality guarantees on k-optimal solutions for
  distributed constraint optimization problems.'' in \emph{IJCAI}, 2007, pp.
  1446--1451.

\bibitem{ZHANG:2005:DSA}
W.~Zhang, G.~Wang, Z.~Xing, and L.~Wittenburg, ``Distributed stochastic search
  and distributed breakout: properties, comparison and applications to
  constraint optimization problems in sensor networks,'' \emph{Artificial
  Intelligence}, vol. 161, no.~1, pp. 55 -- 87, 2005.

\bibitem{Lisy:2010:DCOP_mobile_sensing}
V.~Lisy, R.~Zivan, K.~Sycara, P.~M, V.~Lisy, R.~Zivan, K.~Sycara, and P.~M,
  ``Deception in networks of mobile sensing agents,'' in \emph{9th
  Internationnal Conference on Autonomous Agents and Multiagent Systems (AAMAS
  10), Toronto, Canada}, May 2010, pp. 1031--1038.

\bibitem{Duan:2017:crowdsensing}
Z.~{Duan}, W.~{Li}, and Z.~{Cai}, ``Distributed auctions for task assignment
  and scheduling in mobile crowdsensing systems,'' in \emph{2017 IEEE 37th
  International Conference on Distributed Computing Systems (ICDCS)}, June
  2017, pp. 635--644.

\bibitem{koenig2006power}
S.~Koenig, C.~Tovey, M.~Lagoudakis, V.~Markakis, D.~Kempe, P.~Keskinocak,
  A.~Kleywegt, A.~Meyerson, and S.~Jain, ``The power of sequential single-item
  auctions for agent coordination,'' in \emph{Proceedings of the national
  conference on artificial intelligence}, vol.~21, no.~2.\hskip 1em plus 0.5em
  minus 0.4em\relax Menlo Park, CA; Cambridge, MA; London; AAAI Press; MIT
  Press; 1999, 2006, p. 1625.

\bibitem{notomista:2019:optimal}
G.~Notomista, S.~Mayya, S.~Hutchinson, and M.~Egerstedt, ``An optimal task
  allocation strategy for heterogeneous multi-robot systems,'' in \emph{18th
  European Control Conference (ECC)}.\hskip 1em plus 0.5em minus 0.4em\relax
  IEEE, 2019, pp. 2071--2076.

\bibitem{Prorok:2016:Formalizing}
A.~{Prorok}, M.~A. {Hsieh}, and V.~{Kumar}, ``Formalizing the impact of
  diversity on performance in a heterogeneous swarm of robots,'' in \emph{2016
  IEEE International Conference on Robotics and Automation (ICRA)}, May 2016,
  pp. 5364--5371.

\bibitem{ramach:2019:resilience}
R.~K. Ramachandran, J.~A. Preiss, and G.~S. Sukhatme, ``Resilience by
  reconfiguration: Exploiting heterogeneity in robot teams,'' \emph{arXiv
  preprint arXiv:1903.04856}, 2019.

\bibitem{halasz:2007:dynamic}
A.~Hal{\'a}sz, M.~A. Hsieh, S.~Berman, and V.~Kumar, ``Dynamic redistribution
  of a swarm of robots among multiple sites,'' in \emph{2007 IEEE/RSJ
  international conference on intelligent robots and systems}.\hskip 1em plus
  0.5em minus 0.4em\relax IEEE, 2007, pp. 2320--2325.

\bibitem{BORRMANN:2015:BarrierCertificates}
U.~Borrmann, L.~Wang, A.~D. Ames, and M.~Egerstedt, ``Control barrier
  certificates for safe swarm behavior,'' \emph{IFAC-PapersOnLine}, vol.~48,
  no.~27, pp. 68 -- 73, 2015.

\bibitem{Fisher:1978:submodular}
M.~L. Fisher, G.~L. Nemhauser, and L.~A. Wolsey, \emph{An analysis of
  approximations for maximizing submodular set functions---II}.\hskip 1em plus
  0.5em minus 0.4em\relax Berlin, Heidelberg: Springer Berlin Heidelberg, 1978,
  pp. 73--87.

\bibitem{Ryan:2017:matroid}
R.~K. {Williams}, A.~{Gasparri}, and G.~{Ulivi}, ``Decentralized matroid
  optimization for topology constraints in multi-robot allocation problems,''
  in \emph{2017 IEEE International Conference on Robotics and Automation
  (ICRA)}, May 2017, pp. 293--300.

\bibitem{Shin:2019:sample_greedy}
H.-S. Shin, T.~Li, and P.~Segui-Gasco, ``Sample greedy based task allocation
  for multiple robot systems,'' \emph{arXiv preprint arXiv:1901.03258}, 2019.

\bibitem{Fujishige:2005:submodular}
S.~Fujishige, \emph{Submodular functions and optimization}.\hskip 1em plus
  0.5em minus 0.4em\relax Elsevier, 2005.

\bibitem{luo2020behavior}
W.~Luo, S.~Yi, and K.~Sycara, ``Behavior mixing with minimum global and
  subgroup connectivity maintenance for large-scale multi-robot systems,'' in
  \emph{IEEE International Conference on Robotics and Automation (ICRA)}.\hskip
  1em plus 0.5em minus 0.4em\relax IEEE, 2020, pp. 9845--9851.

\bibitem{luo2019minimum}
W.~Luo and K.~Sycara, ``Minimum k-connectivity maintenance for robust
  multi-robot systems,'' in \emph{IEEE/RSJ International Conference on
  Intelligent Robots and Systems (IROS)}.\hskip 1em plus 0.5em minus
  0.4em\relax IEEE, 2019, pp. 7370--7377.

\bibitem{luo2019voronoi}
------, ``Voronoi-based coverage control with connectivity maintenance for
  robotic sensor networks,'' in \emph{International Symposium on Multi-Robot
  and Multi-Agent Systems (MRS)}.\hskip 1em plus 0.5em minus 0.4em\relax IEEE,
  2019, pp. 148--154.

\bibitem{luo2020multi}
W.~Luo, W.~Sun, and A.~Kapoor, ``Multi-robot collision avoidance under
  uncertainty with probabilistic safety barrier certificates,'' \emph{Advances
  in Neural Information Processing Systems}, vol.~33, 2020.

\bibitem{abramowitz:1988:handbook}
M.~Abramowitz, I.~A. Stegun, and R.~H. Romer, ``Handbook of mathematical
  functions with formulas, graphs, and mathematical tables,'' 1988.

\bibitem{Schlueter:MIDACO}
M.~Schlueter, S.~Erb, M.~Gerdts, S.~Kemble, and J.~Ruckmann, ``Midaco on minlp
  space applications,'' \emph{Advances in Space Research}, vol.~51, no.~7, pp.
  1116--1131, 2013.

\bibitem{Schlueter:ACO}
M.~Schlueter, J.~Egea, and J.~Banga, ``Extended ant colony optimization for
  non-convex mixed integer nonlinear programming,'' \emph{Computers and
  Operations Research}, vol.~36, no.~7, pp. 2217--2229, 2009.

\bibitem{MatlabOTB}
``Matlab optimization toolbox,'' 2018, the MathWorks, Natick, MA, USA.

\end{thebibliography}
\bibliographystyle{IEEEtran}

\end{document}